\documentclass{article}
\usepackage{spconf,amsmath,graphicx,mathtools,amssymb,subfigure,float,hyperref,array,amsfonts,calrsfs,booktabs,multirow,siunitx,etoolbox,xcolor}

\DeclareMathAlphabet{\pazocal}{OMS}{zplm}{m}{n}

\title{Face Landmark-based Speaker-Independent Audio-Visual Speech Enhancement in Multi-Talker Environments}
\name{\begin{tabular}{c}Giovanni Morrone$^{\star}$ \qquad Luca Pasa$^{\dagger}$ \qquad Vadim Tikhanoff $^{\dagger}$ \\ \qquad Sonia Bergamaschi$^{\star}$  \qquad Luciano Fadiga$^{\dagger}$ \qquad Leonardo Badino$^{\dagger}$ \end{tabular}} 
\address{$^{\star}$Department of Engineering "Enzo Ferrari", University of Modena and Reggio Emilia, Modena, Italy \\
$^{\dagger}$Istituto Italiano di Tecnologia, Ferrara, Italy}
\begin{document}
%
\maketitle
\begin{abstract}
In this paper, we address the problem of enhancing the speech of a speaker of interest in a cocktail party scenario when visual information of the speaker of interest is available.

Contrary to most previous studies, we do not learn visual features on the typically small audio-visual datasets, but use an already available face landmark detector (trained on a separate image dataset). 

The landmarks are used by  LSTM-based models to generate time-frequency masks which are applied to the acoustic mixed-speech spectrogram. 
Results show that: \textit{(i)} landmark motion features are very effective  features for this task, \textit{(ii)} similarly to previous work, reconstruction of the target speaker's spectrogram mediated by masking is significantly more accurate than direct spectrogram reconstruction, and \textit{(iii)} the best masks depend on both motion landmark features and the input mixed-speech spectrogram.          

To the best of our knowledge, our proposed models are the first models trained and evaluated on the limited size GRID and TCD-TIMIT datasets, that achieve speaker-independent speech enhancement in a multi-talker setting.

\end{abstract}
\section{Introduction}
\label{sec:intro}
In the context of speech perception, the \textit{cocktail party effect} \cite{cocktail_party, mcdermott} is the ability of the brain to recognize speech in complex and adverse listening conditions where the attended speech is mixed with competing sounds/speech. 

Speech perception studies have shown that watching speaker's face movements could dramatically improve our ability at  recognizing the speech of a target speaker in a multi-talker environment \cite{ZionGolumbic1417, Ma_Wei}.

This work aims at extracting the speech of a target speaker from single channel audio of several people talking simultaneously.
This is an ill-posed problem in that many different hypotheses about what the target speaker says are consistent with the mixture signal.
Yet, it can be solved by exploiting some additional information associated to the speaker of interest and/or by leveraging some prior knowledge about speech signal properties (e.g., \cite{bregman}). 
In this work we use face movements of the target speaker as additional information.

This paper \textit{(i)} proposes the use of face landmark's movements, extracted using Dlib \cite{Kazemi_2014_CVPR, dlib09} and \textit{(ii)} compares different ways of mapping such visual features into time-frequency (T-F) masks, then applied to clean the acoustic mixed-speech spectrogram.

By using Dlib extracted landmarks we relieve our models from the task of learning useful visual features from raw pixels. That aspect is particularly relevant when the training audio-visual datasets are small.

The analysis of landmark-dependent masking strategies is motivated by the fact that speech enhancement mediated by an explicit masking is often more effective than mask-free enhancement \cite{yuxuan_wang_training_2014}.

All our models were trained and evaluated on the GRID \cite{cooke_audio-visual_2006} and TCD-TIMIT \cite{harte_tcd-timit:_2015} datasets in a speaker-independent setting.

\subsection{Related work}
Speech enhancement aims at extracting the voice of a target speaker, while speech separation refers to the problem of separating each sound source in a mixture. Recently proposed audio-only single-channel methods have achieved very promising results \cite{DANet17, Isik2016SingleChannelMS, Kolbaek17}.
However the task still remains challenging. Additionally, audio-only systems need separate models in order to associate the estimated separated audio sources to each speaker, while vision easily allow that in a unified model.

Regarding audio-visual speech enhancement and separation methods an extensive review is provided in \cite{rivet:hal-00990000}. Here we focus on the deep-learning methods that are most related to the present work.

Our first architecture (Section \ref{ssec:vidland2mask}) is inspired by \cite{gabbay_seeing_2017}, where a pre-trained convolutional neural network (CNN) is used to
generate a clean spectrogram from silent video \cite{ephrat2017improved}. Rather than directly computing  a time-frequency (T-F) mask, the mask is computed by thresholding the estimated clean spectrogram. This approach is not very effective since the pre-trained CNN is designed for a different task (video-to-speech synthesis).
In \cite{gabbay2018visual} a CNN is trained to directly estimate clean speech from noisy audio and input video. A similar model is used in \cite{hou_audio-visual_2018}, where the model jointly generates clean speech and input video in a denoising-autoender architecture. 

\cite{hou_audio-visual_2016} shows that using information about lip positions can help to improve speech enhancement. The video feature vector is obtained computing pair-wise distances between any mouth landmarks. 
Similarly to our approach their visual features are not learned on the audio-visual dataset but are provided by a system trained on different dataset. Contrary to our approach, \cite{hou_audio-visual_2016} uses position-based features while we use motion features (of the whole face) that in our experiments turned out to be much more effective than positional features.  

Although the aforementioned audio-visual methods work well, they have only been evaluated in a speaker-dependent setting.
Only the availability of new large and heterogeneous audio-visual datasets has allowed the training of deep neural network-based speaker-independent  speech enhancement models \cite{ephrat_looking_2018, afouras_conversation:_2018, owens2018audio}. 

The present work shows that huge audio-visual datasets are not a necessary requirement for speaker-independent audio-visual speech enhancement. Although we have only considered datasets with simple visual scenarios (i.e., the target speaker is always facing the camera), we expect our methods to perform well in more complex scenarios thanks to the robust landmark extraction.
  
\begin{figure*}[t]
  \centering
  \footnotesize{\quad \textbf{v}: video input \qquad \textbf{y}: noisy spectrogram \qquad$\mathbf{s^m}$: clean spectrogram TBM \qquad\textbf{s}: clean spectrogram IAM \qquad \textbf{m}: TBM \qquad \textbf{p}: IAM}
  
  \subfigure[VL2M]{
    \label{subfig:vl2m}
    \includegraphics[width=.16\textwidth]{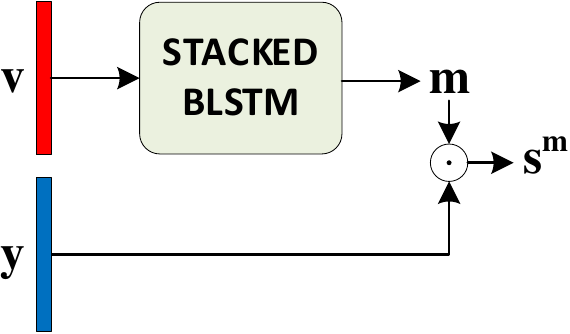}}
  \subfigure[VL2M\_ref]{
    \label{subfig:full_model}
    \includegraphics[width=.32\textwidth]{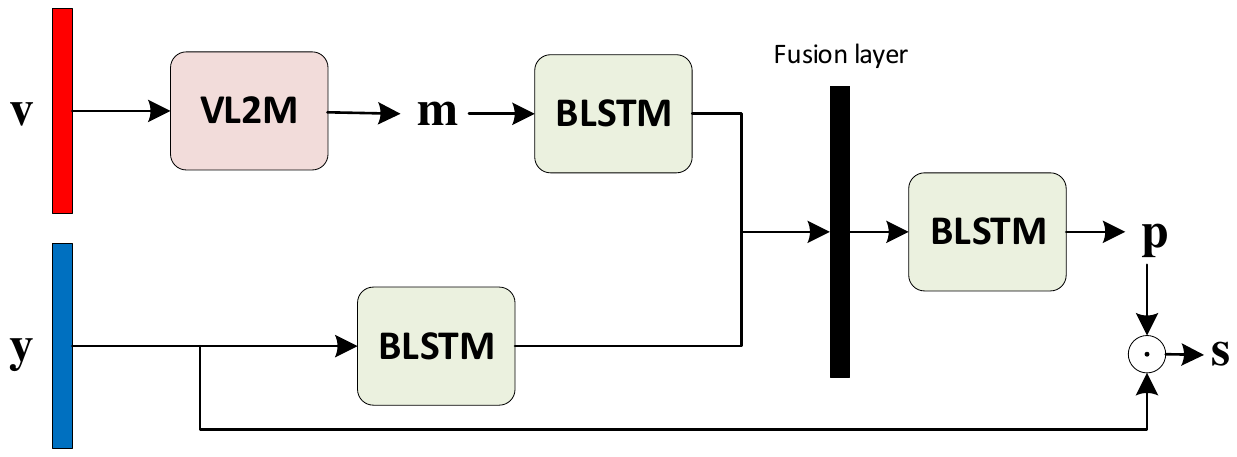}}
  \subfigure[Audio-Visual concat]{
    \label{subfig:concat_model}
    \includegraphics[width=.16\textwidth]{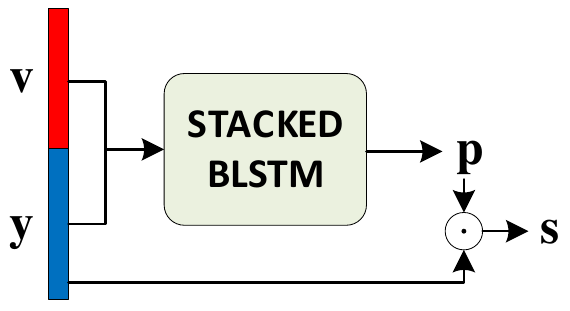}}
  \subfigure[Audio-Visual concat-ref]{
    \label{subfig:concat_ref_model}
    \includegraphics[width=.30\textwidth]{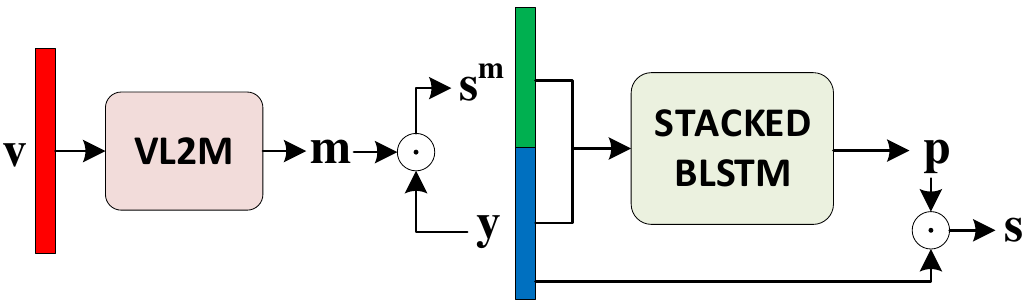}}
  \caption{Model architectures.}
  \label{fig:model}
\end{figure*}

\section{MODEL ARCHITECTURES}
\label{sec:model}
We experimented with the four models shown in Fig. \ref{fig:model}.
All models receive in input the target speaker's landmark motion vectors and the power-law compressed spectrogram of the single-channel mixed-speech signal. All of them perform some kind of masking operation.

\subsection{VL2M model}
\label{ssec:vidland2mask}
At each time frame, the video-landmark to mask (VL2M)  model (Fig. \ref{fig:model}a) estimates a T-F mask from visual features only (of the target speaker).
Formally, given a video sequence $\textbf{v} = [\textbf{v}_1, \dots ,\textbf{v}_T], \, \textbf{v}_t \in \mathbb{R}^n$ and a target mask sequence $\textbf{m} = [\textbf{m}_1, \dots ,\textbf{m}_T], \, \textbf{m}_t \in \mathbb{R}^d$, VL2M perform a function $\mathcal{F}_{vl2m}(\textbf{v}) = \mathbf{\hat{m}}$, where $\mathbf{\hat{m}}$ is the estimated mask. 

The training objective for VL2M is a Target Binary Mask (TBM) \cite{Anzalone2006, Kjems2009}, computed using the spectrogram of the target speaker only. This is motivated by our goal of
extracting the speech of a target speaker as much as possible independently of the concurrent speakers, so that, e.g., we do not need to estimate their number. 
An additional motivations is that the model takes as only input the visual features of the target speaker, and a target TBM that only depends on the target speaker allows VL2M to learn a function (rather than approximating an ill-posed one-to-many mapping).

Given a clean speech spectrogram of a speaker $\mathbf{s}=[\mathbf{s}_1, \dots, \mathbf{s}_T], \, \mathbf{s}_t \in \mathbb{R}^d$, the TBM is defined by comparing, at each frequency bin $f \in [1, \dots, d]$, the target speaker value $\textbf{s}_t[f]$ vs. a reference threshold $\tau[f]$. As in \cite{gabbay_seeing_2017}, we use a function of long-term average speech spectrum (LTASS) as reference threshold. This threshold indicates if a T-F unit is generated by the speaker or refers to silence or noise. The process to compute the speaker's TBM is as follows:
\begin{enumerate}
    \item The mean $\pi[f]$ and the standard deviation $\sigma[f]$ are computed for all frequency bins of all seen spectrograms in speaker's data.
    \item The threshold $\tau[f]$ is defined as $\tau[f] = \pi[f] + 0.6 \cdot \sigma[f]$    where $0.6$ is a value selected by manual inspection of several spectrogram-TBM pairs.
    \item The threshold is applied to every speaker's speech spectrogram $\mathbf{s}$.
    \[	\mathbf{m}_t[f] = \left\{
        \begin{array}{ll}
        1, & \text{if $\mathbf{s}_t[f] \geq \tau[f]$,}  \\
        0, & \text{otherwise.} \\
        \end{array}
        \right. \]
\end{enumerate}

The mapping $\mathcal{F}_{vl2m}(\cdot)$ is carried out by a stacked bi-directional Long Short-Term Memory (BLSTM) network \cite{graves13}.
The BLSTM outputs are then forced to lay within the $[0,1]$ range. Finally the computed TBM $\mathbf{\hat{m}}$ and the noisy spectrogram $\mathbf{y}$ are element-wise multiplied to obtain the estimated clean spectrogram $\mathbf{\hat{s}^m} = \mathbf{\hat{m}} \circ \mathbf{y}$, where $\textbf{y}=[\textbf{y}_1, \dots \textbf{y}_T], \, \textbf{y}_t \in \mathbb{R}^d$.

The model parameters are estimated to minimize the loss:
\begin{equation}
    \scalebox{0.78}[1]{$J_{vl2m} = \sum_{t=1}^T \sum_{f=1}^d -\mathbf{m}_t[f] \cdot \log(\mathbf{\hat{m}}_t[f]) - (1-\mathbf{m}_t[f]) \cdot \log(1-\mathbf{\hat{m}}_t[f])$} \nonumber
\end{equation}

\subsection{VL2M\_ref model}
\label{ssec:full_model}
VL2M generates T-F masks that are independent of the acoustic context. We may want to refine the masking by including such context.
This is what the novel VL2M\_ref does (Fig. \ref{fig:model}b). The computed TBM $\mathbf{\hat{m}}$ and the input spectrogram $\mathbf{y}$ are the input to a function that outputs an Ideal Amplitude Mask (IAM) $\mathbf{p}$ (known as FFT-MASK in \cite{yuxuan_wang_training_2014}). Given the target clean spectrogram $\mathbf{s}$ and the noisy spectrogram $\mathbf{y}$, the IAM is defined as:
\[ \mathbf{p}_t[f] = \frac{\mathbf{s}_t[f]}{\mathbf{y}_t[f]} \]
Note that although IAM generation requires the mixed-speech spectrogram, separate spectrograms for each concurrent speakers are not required.

The target speaker's  spectrogram $\mathbf{s}$ is reconstructed by multiplying the input spectrogram with the estimated IAM. Values greater than $10$ in the IAM are clipped to $10$ in order to obtain better numerical stability as suggested in \cite{yuxuan_wang_training_2014}. 

The model performs a function $\mathcal{F}_{mr}(\textbf{v},\textbf{y}) =  \mathbf{\hat{p}}$ that consists of a VL2M component plus three different BLSTMs $\mathcal{G}_m$, $\mathcal{G}_y$ and $\mathcal{H}$. 

$\mathcal{G}_m(\mathcal{F}_{vl2m}(\textbf{v})) =  \textbf{r}_m$ receives the VL2M mask ${\mathbf{\hat{m}}}$ as input, and $\mathcal{G}_y(\textbf{y}) = \textbf{r}_y$ is fed with the noisy spectrogram.
Their output $\textbf{r}_m, \textbf{r}_y \in \mathbb{R}^z$ are fused in a joint audio-visual representation 
$\mathbf{h}=[\textbf{h}_1, \dots , \textbf{h}_T]$, where $\textbf{h}_t$ is a linear combination of $\textbf{r}_{m_t}$ and $\textbf{r}_{y_t}$: $\mathbf{h}_t = \textbf{W}_{hm} \cdot \textbf{r}_{m_t} + \textbf{W}_{hy} \cdot \textbf{r}_{y_t} + \textbf{b}_h$.
$\mathbf{h}$ is the input of the third BLSTM $\mathcal{H}(\mathbf{h})=\mathbf{\hat{p}}$, where $\mathbf{\hat{p}}$ lays in the [0,10] range.
The loss function is:
\[ J_{mr} =  \sum_{t=1}^T \sum_{f=1}^d (\mathbf{\hat{p}}_t[f] \cdot \mathbf{y}_t[f] -  \mathbf{s}_t[f])^2\]

\subsection{Audio-Visual concat model}
\label{ssub:concat_model}
The third model (Fig. \ref{fig:model}c) performs early fusion of audio-visual features. This model consists of a single stacked BLSTM that computes the IAM mask $\mathbf{\hat{p}}$ from the concatenated $[\mathbf{v},\mathbf{y}]$. The training loss is the same $J_{mr}$ used to train VL2M\_ref. This model can be regarded as a simplification of VL2M\_ref, where the VL2M operation is not performed.

\subsection{Audio-Visual concat-ref model}
\label{ssub:concat_ref_model}
The fourth model (Fig. \ref{fig:model}d) is an improved version of the model described in section \ref{ssub:concat_model}. The only difference is the input of the stacked BLSTM that is replaced by $[\mathbf{\hat{s}^m},\mathbf{y}]$ where $\mathbf{\hat{s}^m}$ is the denoised spectrogram returned by VL2M operation.

\section{Experimental setup}
\subsection{Dataset}
All experiments were carried out using the GRID \cite{cooke_audio-visual_2006} and TCD-TIMIT \cite{harte_tcd-timit:_2015} audio-visual datasets. For each of them, we created a mixed-speech version. 

Regarding the GRID corpus, for each of the $33$ speakers (one had to be discarded) we first randomly selected  $200$ utterances (out of $1000$). Then, for each utterance, we created $3$ different audio-mixed samples. Each audio-mixed sample was created by mixing the chosen utterance with one utterance from a different speaker. 

That resulted in $600$ audio-mixed samples per speaker.

The resulting dataset was split into disjoint sets of $25$/$4$/$4$ speakers for training/validation/testing respectively.

The TCD-TIMIT corpus consists of $59$ speakers (we excluded $3$ professionally-trained lipspeakers) and $98$ utterances per speaker. The mixed-speech version was created following the same procedure as for GRID, with one difference. 
Contrary to GRID, TCD-TIMIT utterances have different duration. Thus $2$ utterances were mixed only if their duration difference did not exceed $2$ seconds. 
For each utterance pair, we forced the non-target speaker's utterance to match the duration of the target speaker utterance. If it was longer, the utterance was cut at its end, whereas if it was shorter, silence samples were equally added at its start and end.

The resulting dataset was split into disjoint sets of $51$/$4$/$4$ speakers for training/validation/testing respectively.

\subsection{LSTM training}
In all experiments, the models were trained using the Adam optimizer \cite{adam}.
Early stopping was applied when the error on the validation set did not decrease over $5$ consecutive epochs.

VL2M, AV concat and AV concat-ref had $5$, $3$ and $3$ stacked BLSTM layers respectively. All BLSTMs had $250$ units. Hyper-parameters selection was performed by using random search with a limited number of samples, therefore all the reported results may improve through a deeper hyper-parameters validation phase.

VL2M\_ref and AV concat-ref training was performed in $2$ steps. We first pre-trained the models using the oracle TBM $\mathbf{m}$. Then we substituted the oracle masks with the VL2M component and retrained the models while freezing the parameters of the VL2M component.

\subsection{Audio pre- and post-processing}
The original waveforms were resampled to 16 kHz. Short-Time Fourier Transform (STFT) $\mathbf{x}$ was computed using FFT size of 512, Hann window of length 25 ms (400 samples), and hop length of 10 ms (160 samples). The input spectrogram was obtained taking the STFT magnitude and performing power-law compression $\mathbf{\lvert x \rvert}^p$ with $p=0.3$. Finally we applied per-speaker 0-mean 1-std normalization.

In the post-processing stage, the enhanced waveform generated by the speech enhancement models was reconstructed by applying the inverse STFT to the estimated clean spectrogram and using the phase of the noisy input signal.

\subsection{Video pre-processing}
Face landmarks were extracted from video using the Dlib \cite{dlib09} implementation of the face landmark estimator described in \cite{Kazemi_2014_CVPR}. It returns 68 x-y points, for an overall 136 values.
We upsampled from 25/29.97 fps (GRID/TCD-TIMIT) to $100$ fps to match the frame rate of the audio spectrogram. Upsampling was carried out through linear interpolation over time. 

The final video feature vector $\mathbf{v}$ was obtained by computing the per-speaker normalized motion vector of the face landmarks by simply subtracting every frame with the previous one. The motion vector of the first frame was set to zero.

\setlength\tabcolsep{6pt}
\begin{table}
  \centering
  \begin{tabular}{lSSS}
    \toprule
    \multirow{2}{*}{} &
      {SDR} & {PESQ} & {ViSQOL} \\
      \midrule
    Noisy & -1.06 & 1.81 & 2.11  \\
    VL2M & 3.17 & 1.51 & 1.16  \\
    VL2M\_ref & $\; \; \mathbf{6.50}$ & $\; \; \mathbf{2.58}$ & $\; \; \mathbf{2.99}$ \\
    AV concat & 6.31 & 2.49 & 2.83 \\
    AV c-ref & 6.17 & $\; \; \mathbf{2.58}$ & 2.96 \\
    \bottomrule
  \end{tabular}
  \caption{GRID results - speaker-dependent. The ``Noisy'' row refers to the metric values of the input mixed-speech signal.}
  \label{tab:grid_spk_dep}
\end{table}
\setlength\tabcolsep{2pt}
\begin{table}
  \centering
  \begin{tabular}{lS @{\hspace{0.5\tabcolsep}}  S  @{\hspace{0.5\tabcolsep}}S|S @{\hspace{0.5\tabcolsep}} S  @{\hspace{0.5\tabcolsep}} S}
    \toprule
    \multirow{2}{*}{} &
      \multicolumn{3}{c|}{2 Speakers} &
      \multicolumn{3}{c}{3 Speakers} \\
      & {SDR} & {PESQ} & {ViSQOL} & {SDR} & {PESQ} & {ViSQOL} \\
      \midrule
    Noisy & 0.21 & 1.94 & 2.58 & -5.34 & 1.43 & 1.62 \\
    VL2M & 3.02 & 1.81 & 1.70 & -2.03 & 1.43 & 1.25 \\
    VL2M\_ref & 6.52 & 2.53 & 3.02 & 2.83 & 2.19 & 2.53 \\
    AV concat & 7.37 & 2.65  & 3.03 & 3.02 & 2.24 & 2.49 \\
    AV c-ref & $ \; \; \mathbf{8.05}$ & $ \; \; \mathbf{2.70}$ & $ \; \; \mathbf{3.07}$ & $ \; \; \mathbf{4.02}$ & $ \; \; \mathbf{2.33}$ & $ \; \; \mathbf{2.64}$\\
    \bottomrule
  \end{tabular}
  \caption{GRID results - speaker-independent.}
  \label{tab:grid}
\end{table}
\begin{table}
  \centering
  \begin{tabular}{lS @{\hspace{0.5\tabcolsep}}  S  @{\hspace{0.5\tabcolsep}}S|S @{\hspace{0.5\tabcolsep}} S  @{\hspace{0.5\tabcolsep}} S}
    \toprule
    \multirow{2}{*}{} &
      \multicolumn{3}{c|}{2 Speakers} &
      \multicolumn{3}{c}{3 Speakers} \\
      & {SDR} & {PESQ} & {ViSQOL} & {SDR} & {PESQ} & {ViSQOL} \\
      \midrule
    Noisy & 0.21 & 2.22 & 2.74 & -3.42 & 1.92 & 2.04 \\
    VL2M & 2.88 & 2.25 & 2.62 & -0.51 & 1.99 & 1.98 \\
    VL2M\_ref & 9.24 & 2.81 & 3.09 & 5.27 & 2.44 &  2.54 \\
    AV concat & 9.56 & 2.80 & 3.09 & 5.15 & 2.41 & 2.52 \\
    AV c-ref & $ \; \; \mathbf{10.55}$ & $ \; \; \mathbf{3.03}$ & $ \; \; \mathbf{3.21}$ & $ \; \; \mathbf{5.37}$ & $ \; \; \mathbf{2.45}$ & $ \; \; \mathbf{2.58}$ \\
    \bottomrule
  \end{tabular}
  \caption{TCD-TIMIT results - speaker-independent.}
  \label{tab:tcdtimit}
\end{table}

\section{Results}
In order to compare our models to previous works in both speech enhancement and separation, we evaluated the performance of the proposed models using both speech separation and enhancement metrics. Specifically, we measured the capability of separating the target utterance from the concurrent utterance with the source-to-distortion ratio (SDR) \cite{vincent_performance_2006, raffel2014mir_eval}. While the quality of estimated target speech was measured with the perceptual PESQ \cite{rix_perceptual_2001} and ViSQOL \cite{hines_visqol:_2012} metrics. For PESQ we used the narrow band mode while for ViSQOL we used the wide band mode.

As a very first experiment we compared landmark position vs. landmark motion vectors. It turned out that landmark positions performed poorly, thus all results reported here refer to landmark motion vectors only. 

We then carried out some speaker-dependent experiments to compare our models with previous studies as, to the best of our knowledge, there are no reported results of speaker-independent systems trained and tested on GRID and TCD-TIMIT to compare with.
Table \ref{tab:grid_spk_dep} reports the test-set evaluation of speaker-dependent models on the GRID corpus with landmark motion vectors. Results are comparable with previous state-of-the-art studies in an almost identical setting \cite{gabbay_seeing_2017, gabbay2018visual}.

Table \ref{tab:grid} and \ref{tab:tcdtimit} show speaker-independent test-set results on the GRID and TCD-TIMIT datasets respectively. V2ML performs significantly worse than the other three models indicating that a successful mask generation has to depend on the acoustic context.
The performance of the three models in the speaker-independent setting is comparable to that in the speaker-dependent setting.
 
AV concat-ref outperforms V2ML\_ref and AV concat for both datasets. This supports the utility of a refinement strategy and suggests that the refinement is more effective when it directly refines the estimated clean spectrogram, rather than refining the estimated mask.

Finally, we  evaluated the systems in a more challenging testing condition where the target utterance was mixed with $2$ utterances from $2$ competing speakers. 
Despite the model was trained with mixtures of two speakers, the decrease of performance was not dramatic. 

Code and some testing examples of our models are available at \url{https://goo.gl/3h1NgE}.

\section{Conclusion}
This paper proposes the use of face landmark motion vectors for audio-visual speech enhancement in a single-channel multi-talker scenario. Different models are tested where landmark motion vectors are used to generate time-frequency (T-F) masks that extract the target speaker's spectrogram from the acoustic mixed-speech spectrogram.

To the best of our knowledge, some of the proposed models are the first models trained and evaluated on the limited size GRID and TCD-TIMIT datasets that accomplish speaker-independent speech enhancement in the multi-talker setting,  with a quality of enhancement comparable to that achieved in a speaker-dependent setting.


\bibliographystyle{IEEEbib}
\bibliography{references}
\end{document}